\documentclass{template}

\usepackage{tikz}
\usepackage{subcaption}
\definecolor{myLightGray}{RGB}{191,191,191}
\definecolor{myGray}{RGB}{160,160,160}
\usetikzlibrary{patterns}
\usetikzlibrary{decorations.pathreplacing}
\usepackage{mathtools}

\DeclarePairedDelimiter\floor{\lfloor}{\rfloor}

\begin{document}

\title{Predicting the probability distribution of bus travel time to move towards reliable planning of public transport services}

\author{L\'ea Ricard}
\affil{Department of Computer Science, Universit\'e de Montr\'eal, Montr\'eal, Canada}
\author{Guy Desaulniers}
\author{Andrea Lodi}
\author{Louis-Martin Rousseau}
\affil{Department of Mathematics and Industrial Engineering, Polytechnique Montr\'eal, Montr\'eal, Canada}

\date{}

\maketitle

\begin{abstract}
An important aspect of the quality of a public transport service is its reliability, which is defined as the invariability of the service attributes. Preventive measures taken during planning can reduce risks of unreliability throughout operations. In order to tackle reliability during the service planning phase, a key piece of information is the long-term prediction of the density of the travel time, which conveys the uncertainty of travel times. We introduce a reliable approach to one of the problems of service planning in public transport, namely the Multiple Depot Vehicle Scheduling Problem (MDVSP), which takes as input a set of trips and the probability density function (p.d.f.) of the travel time of each trip in order to output delay-tolerant vehicle schedules. This work empirically compares probabilistic models for the prediction of the conditional p.d.f. of the travel time, as a first step towards reliable MDVSP solutions. Two types of probabilistic models, namely similarity-based density estimation models and a smoothed Logistic Regression for probabilistic classification model, are compared on a dataset of more than 41,000 trips and 50 bus routes of the city of Montréal. The result of a vast majority of probabilistic models outperforms that of a Random Forests model, which is not inherently probabilistic, thus highlighting the added value of modeling the conditional p.d.f. of the travel time with probabilistic models. A similarity-based density estimation model using a $k$ Nearest Neighbors method and a Kernel Density Estimation predicted the best estimate of the true conditional p.d.f. on this dataset.  
\end{abstract}

\section{Introduction}
In order to increase the ridership and attract new users, public transport agencies put increasing emphasis on improving the quality of the service they provide and particularly its regularity, also referred to as reliability \citep{Ma2014}. Studies show that a majority of passengers put more value on a reduction of the travel time (TT) variability than on a reduction of TT itself \citep{Bates2001}. However, day-to-day disruptions and delays are considered unavoidable on the day of operation \citep{Amberg2019,Kramkowski2009} due to the randomness of incidents and to the variation in demand and capacity factors. Bus sharing the road with other road-based vehicles (e.g., cars, bikes and trucks) are likely to have even higher degrees of variability, because they are subject to the same - morning and evening peaks - traffic patterns \citep{Comi2017}. 

TT variability is explained by yearly, monthly, day-to-day and hourly variability as well as vehicle-to-vehicle variability \citep{Bachu2014, Buchel2018,Kieu2014}. Daily variability is the fluctuation between the TT of a trip at a specific time on different days whereas hourly variation is the variation of a route's TT during the course of the day. A bus route is defined by an ordered sequence of road segments and bus stops, where the first and the last stops are called terminals. A bus line usually has two associated routes, each one going in opposite directions (e.g., North-South or East-West axis). A trip is defined as a unique travel which follows a specific route according to a predefined schedule. TTs are usually longer during peak periods due to denser traffic and/or passenger flows. Indeed, dwell time increases with the number of passengers, because the loading and unloading are more laborious \citep{Buchel2018}. Bus driver behavior and vehicle type (e.g., standard or articulated) are examples of underlying causes of vehicle-to-vehicle variability. 

Reliability can be addressed prior to the operational phase. At the strategic planning level, adding reserved lanes for buses can increase the reliability of the service, while during operations, bus holding is a popular solution to alleviate risks of bus bunching. The latter consists in holding a bus at key locations along the route if it is running ahead of time. However, service reliability is rarely taken into account at the tactical planning level, when the detailed planning of the service is computed \citep{Oort2011}. Service planning in public transport is usually divided in five problems that can be addressed in a sequential or (partially) integrated order: network design, headway setting between two consecutive trips along a same route or timetabling setting, vehicle scheduling, duty scheduling and crew rostering. The second problem specifies for a set of trips their departure times from terminals and all subsequent stops. The vehicle scheduling problem assigns vehicles to cover the latter set of trips, in such a way that every trip is covered exactly once and at minimal costs. When the operator's fleet is spread in two or more depots, it is referred to as the Multiple Depot Vehicle Scheduling Problem (MDVSP). Duty scheduling consists in building anonymous duties, which are sequences of trip segments and breaks forming a driver workday and respecting various working rules, to cover all trips to operate on a given day. Finally, crew rostering assigns those duties to the drivers over a one-week or one-month horizon. 

One of the key information for all five problems or one in particular (vehicle scheduling) is the TT prediction. There are two types of TT prediction: short-term and long-term. Both types can predict either a segment or a complete trip TT. The former is usually performed less than one hour before a trip and uses online information as well as external factors (e.g., weather). This type of prediction can be integrated to the operator's operations control system and provides online information to the users about the estimated arrival time of a bus. On the other hand, the long-term TT prediction can be performed a few days before the trip and helps for transit planning.

In this work, we explore methods for the long-term prediction of public bus TT in order to address reliability when solving the MDVSP. We define the reliable MDVSP with stochastic TTs, which takes as input a set of timetabled trips and the conditional probability density function (p.d.f.) of their TT in order to output cost-efficient and delay tolerant vehicle schedules. We frame the long-term prediction of the density of the TT (PDTT) as a supervised learning problem which aims at predicting, for each trip in a set of timetabled trips, the complete conditional p.d.f. of its TT where the condition corresponds to the trips' characteristics. The objective of this problem is to predict the complete distribution of the TT instead of the expected TT, in order to capture the inherent uncertainty associated with the duration of bus trips. Models that fit the conditional p.d.f. of the TT are compared in order to select the one that outputs the best estimate of the true conditional p.d.f. of the TT given the feature vector of a trip. Probabilistic models are compared to a Random Forest model, which provided the most promising results among the three regression models studied in the work of \cite{Mendes-Moreira2012}. We introduce two types of probabilistic models, namely the similarity-based density estimation models and the smoothed logistic regression model for probabilistic classification, and present experimental results on a large-scale dataset of more than 41,000 trips and 50 bus routes. Our contribution is two-fold: 
\begin{itemize}
    \item The state-of-the-art for long-term prediction of public bus TT is almost nonexistent \citep{Moreira-Matias2015}. This work tries to fill this gap and, in addition, it is to our knowledge the first work to propose probabilistic models for the long-term prediction of public bus TT.
    \item To the best of our knowledge, it is the first study in the field of public transport that empirically studies such a large number of bus route's TTs simultaneously. We hope this improves the transferability of our results to other bus networks.
\end{itemize}

The remainder of this paper is organized as follows. Section 2 defines the reliable MDVSP with stochastic TTs and sets out its subproblem, the PDTT, which consists in predicting the long-term conditional p.d.f. of the TT. In Section 3, we review the literature on TT analysis. The dataset used and its features are discussed in Section 4. Section 5 describes the methodology that can be applied for the PDTT. In Section 6, data preparation as well as features and parameters selection are presented, before comparing the performance of all models for the prediction of the p.d.f. of the TT. Section 7 summarizes our findings. 

\section{The reliable MDVSP with stochastic TTs}
\label{sec:reliable}
The MDVSP outputs a set of vehicle schedules, which are sequences of trips and waiting times starting and ending at the same depot, such that each travel is either a deadhead trip (e.g., between a depot and a terminal or between two terminals) or a timetabled trip. A deadhead trip between two terminals enables the connection of two timetabled trips ending and starting at different terminals. The reliability of a vehicle schedule is often assessed by its delay tolerance, which is defined as the aptitude to prevent delays to propagate to timetabled trips downstream. This type of delay is called secondary delay, whereas deviation from the planned duration of a timetabled trip caused by a disruption or variability during operation is called primary delay. 

The reliable MDVSP with stochastic TTs takes the set $\mathcal{V}$ of $n$ timetabled trips and the long-term prediction of the TT of each of these trips in input.  Let $D$ be the set of depots, $S$ the set of all feasible vehicle schedules, and $S^d$ the subset of schedules starting and ending at depot $d$. The problem is to find a subset of vehicle schedules in $S$ that covers exactly once each timetabled trip while respecting the number of available buses $b_d$ at each depot $d\in D$ and minimizing a weighted sum of the total planned vehicle operating cost and the total expected secondary delay. To formulate this problem, we define for each trip $i\in V$ and schedule $s\in S$ a binary parameter $a_{is}$ which is equal to 1 if schedule $s$ covers trip $i$ and 0 otherwise, and denote by $c_s$ the cost of schedule $s$ (including expected delay penalties). Furthermore, we introduce for each schedule $s\in S$, a binary variable $y_s$ that takes value 1 if schedule $s$ is selected in the solution and 0 otherwise. The reliable MDVSP with stochastic TTs can then be expressed as the following integer linear program: 

\begin{align}
    \label{eq:SP1}
    \text{min \qquad}&\sum_{s \in \mathcal{S}} c_s y_s \\
    \label{eq:SP2}
    \text{s.t. \qquad}&\sum_{s \in \mathcal{S}} a_{is}y_s = 1, \forall i \in \mathcal{V} \\
    \label{eq:SP3}
    &\sum_{s \in \mathcal{S}^d}  y_s \leq b_d, \forall d \in \mathcal{D}\\
    \label{eq:SP4}
    &y_s \in \{0,1\}, \forall s \in \mathcal{S}.
\end{align}

Constraints (\ref{eq:SP2}) ensure that each timetabled trip is covered by a selected vehicle schedule, whereas constraints (\ref{eq:SP3}) impose vehicle availability at each depot.

The objective function (\ref{eq:SP1})  minimizes the total cost of the selected schedules which combines planned operational costs and expected secondary delay penalties. The planned costs usually include a fix cost per vehicle used and a variable cost that depends on the traveled distance and the waiting time attended by a bus driver. Consider a vehicle schedule $s = \{v_1, v_2, ..., v_{m_s}\} \in \mathcal{S}$ of $m_s$ timetabled trips. For notational conciseness, we denote trip $v_i$ by $i$ directly in the following. The total cost $c_s$ of the vehicle schedule $s$ is a weighed sum of the planned costs $q_s$ and the sum of the expected secondary delay $\mathbb{E}(R_i)$ of all trips $i$ covered by the schedule $s$, weighted by a factor $\beta$:

\begin{equation}
\label{cost}
    c_s = q_s + (\beta \sum_{i=1}^{m_s} \mathbb{E}(R_i)).
\end{equation}

The secondary delay $R_i$ of a trip $i$ is the difference between its actual departure time $D_i$ and the planned departure time $d_i$, computed as

\begin{equation}
    R_i = D_i - d_i.
\label{eq:sed_delay}
\end{equation}

The random variable $D_i$ is a convolution of the previous trip's actual departure time ($D_{i-1}$), actual TT ($T_{i-1}$) and the minimum in-between time between trips $i$ and $i-1$ ($l_{i-1,i}$):

\begin{align}
    &D_i = max\{D_{i-1} + T_{i-1} + l_{i-1,i}, d_i\}\\
    &D_1 = d_1.
\label{eq:rec_dep}
\end{align}

Precisely, $l_{i-1,i}$ accounts for the deadhead travel between terminals, if the trip $i-1$ ends at a different terminal than the departure terminal of trip $i$, and the minimum break time for drivers.

The expected value of $R_i$ is computed by integrating over the p.d.f. of the secondary delay of trip $i$. Indeed, the expected value of a continuous random variable $X$ with a p.d.f. $p(\cdot)$ is $\mathbb{E}(X) = \int xp(x)dx$. We have that the p.d.f. of $R_i$ is recursively computed using the p.d.f. of $T_1$ to $T_{i-1}$. Indeed, the expected values of $T_1$ to $T_{i-1}$ are not sufficient to compute the expected value of $R_i$. Thus, instead of being interested in the long-term prediction of the expected TT of a set of trips, we will rather focus on the long-term prediction of the p.d.f. of the TT of a set of trips. We call the latter approach the PDTT.

We want to learn a predictive model for the long-term p.d.f. of the TT in order to minimize the total expected secondary delays when solving the reliable MDVSP with stochastic TTs.  The PDTT computes for each trip $i$ in a set of unseen timetabled trips (test set), an estimate of its conditional p.d.f., $\hat{p}(T_i \mid \boldsymbol{x}_i)$, given $\boldsymbol{x}_i$ the set of characteristics of trip $i$. It is assumed that the random variables $\{T_1, \dots, T_n\}$ are independent and identically distributed. The estimator $\hat{p}(\cdot)$ is learned from observed trips in a training set, such that $\hat{p}(\cdot)$ is fitted in order to minimize the negative log-likelihood (NLL) of the training set (containing $n_{train}$ points), computed as

\begin{align}
    NLL_{train} = - \sum_{i = 1} ^{n_{train}} \log( \hat{p}(t_{i}|\boldsymbol{x}_{i})), 
\label{eq:NLPD}
\end{align}

\noindent with $t_{i}$ the true TT of trip $i$. If possible, the performance of the PDTT should be evaluated by the test mean square error (MSE) of the expected secondary delay

\begin{align}
     MSE = \frac{1}{n_{test}}\sum_{i=1}^{n_{test}} \left(r_i - \mathbb{E}(\hat{R}_i)\right)^2,
\end{align}

\noindent with $n_{test}$ is the number of test points, $\mathbb{E}(\hat{R}_i)$ the estimated value of the expected secondary delay computed using $\hat{p}(T_i \mid \boldsymbol{x}_i)$ and $r_i$ the true secondary delay of trip $i$. Indeed, the purpose of the PDTT is to improve the reliability of the MDVSP solutions and this starts by accurate expected secondary delays.  We will fit PDTT models on a sample of the observed timetabled trips, namely the ones of the routes with the largest number of recorded trips (see Section \ref{sec:data}). Since this dataset was collected from real buses circulating in the city of Montréal, many actual vehicle schedules contain timetabled trips from routes not in the sample studied. For such vehicle schedules, it is not possible to compute $D_i$ and $R_i$ for most of the trips $i$ in the schedule. It is thus impossible to evaluate the PDTT models in terms of the expected secondary delay error of all trips in the test set in this work. For that reason, the performance of the PDTT will be evaluated in terms of the NLL score as well, but of the test set. Generally, if a PDTT model achieves a better test NLL, it also achieves a lower expected secondary delay error. The NLL of the test set is analogue to the one of the training set and is computed as

\begin{align}
    NLL_{test} = - \sum_{i = 1} ^{n_{test}} \log( \hat{p}(t_{i}|\boldsymbol{x}_{i})).
\label{eq:NLPD_test}
\end{align}

To summarize, a PDTT model outputs for every trip $i$ in a set of trips an estimate of the conditional p.d.f. of its TT, $\hat{p}(T_i \mid \boldsymbol{x}_i)$. This differs from \textit{point prediction} models \citep{dutordoir2018}, as here we are interested in predicting the complete probability distribution instead of only the expected value of the TT. This will enable us to compute and minimize the expected secondary delays of a set of timetabled trips when solving the reliable MDVSP with stochastic TTs. 

\section{Related works}
\label{related_works}

The introduction of automatic vehicle location (AVL) data has given rise to a flourishing number of studies in the field of public transport on speed, arrival time and TT analysis. Because TT and arrival time measures are closely related, studies on both measures are treated without distinction. Indeed, the arrival time $A_i$ of a trip $i$ is given by

\begin{equation}
    A_i = D_{i} + T_i.
\end{equation}

In this section, three topics are covered: long-term TT prediction, TT variability analysis and TT distribution modeling. The latter fits the TT distribution of trips that occurred during a given period in order to analyze the shape and nature of the p.d.f. of the TT, without trying to predict future events, as in the PDTT.  We extend the field of the first topic to all road-based transport, but the subsequent topics are restricted to the public transport field. Approaches proposed for the PDTT are inspired by lessons learned through the review of the literature on these topics.

Compared with the literature on short-term TT prediction, studies on long-term TT prediction are rare and to the best of our knowledge, only the works of \citet{Chen2018}, \citet{Mendes-Moreira2012} and \citet{Klunder2007} proposed or reviewed long-term TT prediction methods. In a survey on improving the planning of public transit using AVL data, \citet{Moreira-Matias2015} suggested that the long-term TT prediction should be valid for an horizon of at least the entire forecasting period. They divided models found in the literature for short-term TT prediction in four categories: (i) machine learning and regression, (ii) state-based and time-series, (iii) traffic theory-based and (iv) historical databased, and suggested that some regression algorithms applied for short-term TT prediction could be adapted for long-term TT prediction. Gradient boosting is an example of a model that was successfully applied to the short-term TT prediction and then adapted by \citet{Chen2018} to the long-term TT prediction of trips on a freeway segment in Taiwan. The features were ranked in order of relative importance: time of the day, day of the week, national holiday, day of long / consecutive holiday, big event / activity, electronic tool collection fee promotion and narrowing of roadway. Three regression models, namely a Projection Pursuit Regression, a Support Vector Machine and a Random Forest, were compared by \citet{Mendes-Moreira2012} for the long-term TT prediction of trips of one public bus route in Porto. With a basic pre-processing work, the Random Forest had better results, but was slightly outperformed by a Projection Pursuit Regression when the authors added an instance selection step. \citet{Klunder2007} trained a $k$ Nearest Neighbours algorithm ($k$NN) with only time-based variables for the long-term TT prediction on a motorway network in the Netherlands.

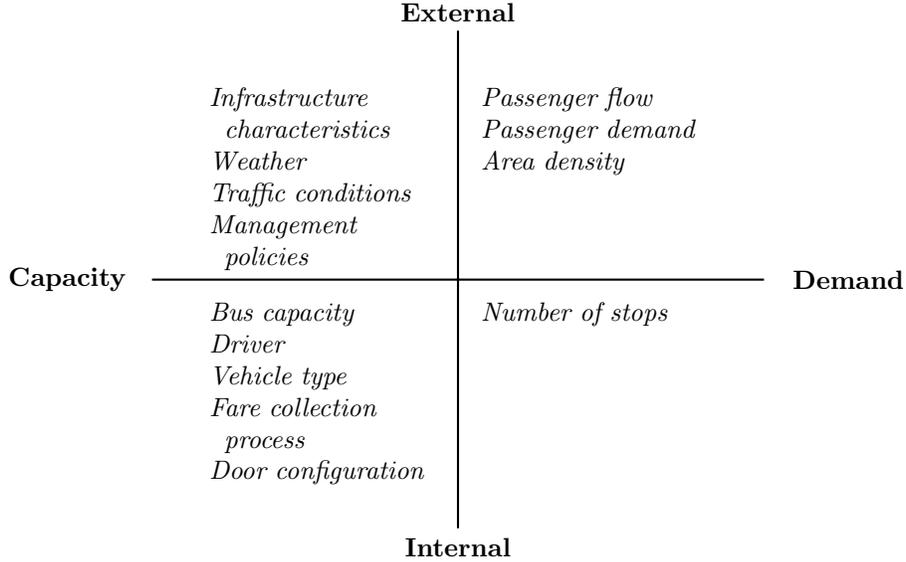
\begin{figure}[!ht]    
\centering
\begin{tikzpicture}[]
  \begin{scope}[every node/.style={%
  text width=3cm,text depth=2cm,inner sep = 3mm,
  minimum height=2cm,minimum width=2.5cm,
  align=left}]

\node[](T1){%
    \textit{Infrastructure \\ \hspace{0.2cm}characteristics\\Weather\\Traffic conditions\\Management \\ \hspace{0.2cm}policies}};

 \node[anchor=north west](T2) at (T1.north east){%
 \textit{Passenger flow\\Passenger demand\\Area density}};

\node[anchor=north west](T3) at (T1.south west){%
\textit{Bus capacity\\Driver\\Vehicle type\\Fare collection\\ \hspace{0.2cm}process\\Door configuration}};

\node[anchor=north west](T4) at (T1.south east){%
\textit{Number of stops}}; 
\end{scope}

\begin{scope}[align=center]
 \node [text width=6cm,above=3ex,anchor=south] (T5) at (T1.north east)
   {\textbf{External}};

 \node [text width=6cm,below=3ex,anchor=north] (T6) at (T3.south east) {%
    \textbf{Internal}}; 
    
\node [text width=2cm,left=3ex,anchor=east](T7) at (T1.south west) 
   {\textbf{Capacity}};

 \node [text width=2cm,right=3ex,anchor=west](T8) at (T2.south east) {%
   \textbf{Demand}};
\end{scope}

\draw[thick] (T5) -- (T6)
             (T7) -- (T8) ;   
\end{tikzpicture} 
\caption{Reasons of TT variability}
  \label{fig:causes_variance}
\end{figure}

The reasons for TT variability can be external or internal \citep{Yetiskul2012} and related to demand or capacity \citep{Mazloumi2009} (see Figure \ref{fig:causes_variance}). In an early study, \citet{Abkowitz1983} suggested that shorter routes may have reduced TT variablity. Also, they reported that a running time deviation at the beginning of a route tends to propagate downstream. Hence, control actions to correct early deviations on a route could reduce TT variability. \citet{Strathman1993} reported that the afternoon peak period has higher TT variability, in particular because of the higher passenger demand.  In a study on TT variability in the city of Ankara, \citet{Yetiskul2012} found major differences in regional TT variability and suggested that bus-stop spacing should depend on the neighborhood density. \citet{Comi2017} performed a time series decomposition of the TT and compared it to the temporal traffic patterns. The two are reported to have similarities. Also, the seasonality of the time series decomposition was most significant for the hour of the day.

TT distribution modeling has been studied mostly with the objective of quantifying the reliability of a transit service. Most works on TT distribution modeling in public transit occurred after the introduction of AVL systems. We focus our review on the work of \citet{Mazloumi2009}, \citet{Ma2016} and \citet{Buchel2018}, as they are, in our view, the most comprehensive studies, from which useful lessons can be learned for the PDTT. \citet{Mazloumi2009} assessed the shape and nature of the TT distribution over the course of the day and for different levels of temporal aggregation on segments of a bus route. The authors measured the level of temporal aggregation by the length of the departure time windows (DTW), which are time slots for which trips departing during the slot are aggregated for subsequent analyzes (e.g., 15 minutes, 30 minutes or 1 hour). The study concluded that for shorter DTWs, the TT distribution follows a Normal distribution. For longer DTWs, this result holds for peak periods, but not for off-peak periods. For the latter, the Log-Normal distribution fits better. Also, the contribution of a set of features to the TT was assessed  through a linear regression analysis. The land use (industrial vs. residential) and the length of the segment were those affecting the most the TT variance. \cite{Ma2016} studied intensively the influence of temporal and spatial aggregation on the TT distribution, with the objective of providing common grounds for modeling and evaluating the performance. To this end, several settings of temporal and spatial aggregations were assessed and an evaluation approach based on a statistical hypothesis test was proposed. The TT of a trip starting at stop $i$ and ending at stop $j$ was decomposed in dwelling times $DT_k$ at each stop $k$ of the trip and running times $RT_{(k,k+1)}$ between each pair of stops

\begin{equation}
    T_{(i,j)} = \sum_{k = i}^{j - 1}DT_k + RT_{(k,k+1)}.
\end{equation}

Results concerning the normality of the TT distribution were in line with the ones of \cite{Mazloumi2009}. Also, the analysis suggested that spatial aggregation tends to decrease the multimodality of TT distribution. A multimodal distribution is defined as a probability distribution with several modes. A Gaussian mixture model (GMM) was proposed to address the multimodality of the link level TT distribution. \cite{Buchel2018} found that the Log-Normal distribution was, out of four unimodal statistical distributions, the best fit for the TT distribution modeling.

\section{Data}
\label{sec:data}
\begin{figure}[!ht]
    \centering
    \includegraphics[scale=0.5]{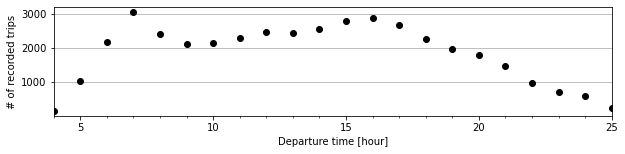}
    \caption{Total number of recorded trips per hour in the dataset}
    \label{fig:nb_recorded_trips}
\end{figure}

The dataset used for this study was collected during a 2-month period from  08/28/2017 to 10/29/2017 by in-car Advanced Public Transport Systems (APTS) installed in buses running in the city of Montréal, Canada. Those systems collect automatically at every stop of a trip the corresponding trip identifier, route identifier, direction identifier, stop identifier, date, scheduled departure time, scheduled arrival time, actual departure time, actual arrival time and number of passengers loading or unloading, among other things. The scheduled departure, scheduled arrival, actual departure and actual arrival times are stored in milliseconds. The actual TT of a trip is the difference between the actual arrival time and the actual departure time at the terminals. Hence, for every route, only the first and last stops (i.e., terminals) data is kept.  Since the APTS were embedded in approximately  20\% to 30\% of the vehicles at that time, weekends and holidays had an insufficient number of trips recorded. Indeed, during weekends and holidays, the service is reduced and thus the number of trips recorded during those days is too small to conduct relevant data-driven analysis. For that reason, weekends and holidays are not studied and are removed from the dataset. After removing weekends and holidays, the dataset has more than 116,000 trips. Of the 408 routes in the dataset, only the 50 most frequent are kept for the remainder of the study, resulting in a dataset of over 41,000 trips. Figure \ref{fig:nb_recorded_trips} shows that the 50 selected routes run between 4:00AM to 1:59AM (+1 day) during weekdays. To facilitate the notation, we add two extra hours to the usual 24-hour daily period. Thus, we say that the selected routes run between 4:00AM to 25:59PM.

\subsection{Route's characteristics}

\begin{figure}[!ht]
    \centering
    \includegraphics[scale=0.5]{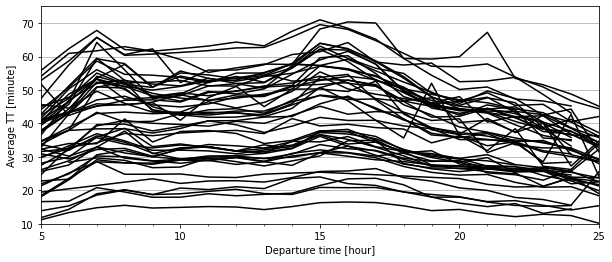}
    \caption{Average TT per hour}
    \label{fig:hourly_TT}
\end{figure}

The distribution of the average TT per hour is presented in Figure \ref{fig:hourly_TT}, where each broken line represents the evolution of the average TT of a route. It is possible to distinguish the morning peak approximately from 6:00AM to 8:59AM, and the afternoon peak, approximately from 14:00PM to 17:59PM for all routes. As for the average TT, the number of recorded trips is also higher during peak periods (see Figure \ref{fig:nb_recorded_trips}). The afternoon peak usually has an average TT higher than the morning peak and the average TT as well as the number of recorded trips per hour are generally decreasing after 17:00PM.

The main characteristics of the 50 selected routes, namely the number of stops, the distance traveled and the type of operational region, are presented in Table \ref{table:routes_characteristics}. Each route has a unique combination of line identifier and direction, such that \textit{A-East} and \textit{A-West} are two different routes of the same bus line, but in opposite directions. We categorized the type of operational region in 6 categories: residential areas, crossing city center (CC), from city center (to a residential area), to city center (from a residential area), from an industrial (indust.) area (to a residential area), to an industrial area (from a residential area). Bus line B is the only one crossing the city center; it starts in a residential neighborhood, crosses the city center and ends in another residential neighborhood. Industrial areas are characterized by a high density of factories. The  city center and industrial areas are usually regions where a large number of people commute to work every day. The majority of bus routes operate in residential areas (32 out of 50). Those bus routes may, for example, connect two residential areas or a residential area to a subway or train station. The number of stops per bus route ranges from 17 to 74 stops, while the distance traveled ranges from 3.0 km to 15.3 km. In general, as the number of stops goes up, the distance traveled goes up as well. Line P is a counterexample, because it has a large number of stops close to each other.  Note that the number of stops and the distance traveled of two routes of the same line are generally not equal, as the path in one direction is usually not symmetric to the path in the other direction (e.g., because some streets are one-way). 

\begin{table}[ht!]
\caption {Characteristics of bus routes studied}
\label{table:routes_characteristics}
\begin{center}
\begin{tabular}{lllll|lllll}
    \toprule
    Line&Dir.&\#stops&Dist.&Type of&Line&Dir.&\#stops&Dist.&Type of\\
    &&&(km)&Region&&&&(km)&Region\\
    \midrule
    A&East&52&13.9&Residential&M&West&18&4.0&Residential\\
    A&West&49&14.5&Residential&N&East&28&5.3&Residential\\
    B&East&46&13.2&Cross CC&N&West&29&5.3&Residential\\
    B&West&46&12.0&Cross CC&O&North&35&3.0&To indust.\\
    C&North&40&12.1&Residential&O&South&40&3.4&From indust.\\
    C&South&45&12.3&Residential&P&East&74&9.0&Residential\\
    D&North&33&10.3&From indust.&P&West&67&8.5&Residential\\
    D&South&36&10.4&To indust.&Q&East&40&7.0&Residential\\
    E&East&50&14.2&Residential&Q&West&38&7.0&Residential\\
    E&West&52&13.3&Residential&R&East&37&5.3&Residential\\
    F&East&34&7.8&Residential&R&West&35&5.3&Residential\\
    F&West&36&7.7&Residential&S&East&47&11.8&From indust.\\
    G&North&17&4.6&Residential&S&West&51&11.6&To indust.\\
    G&South&19&4.3&Residential&T&North&34&8.5&To indust.\\
    H&North&37&9.3&Residential&T&South&30&8.5&From indust.\\
    H&South&40&10.8&Residential&U&North&46&11.1&Residential\\
    I&East&71&15.3&Residential&U&South&42&10.7&Residential\\
    I&West&68&15.3&Residential&V&East&46&9.5&Residential\\
    J&North&28&7.1&From CC.&V&West&49&8.5&Residential\\
    J&South&30&7.1&To CC&W&East&43&10.3&Residential\\
    K&East&53&11.1&To CC&W&West&47&11.6&Residential\\
    K&West&51&11.4&From CC&X&North&30&6.6&From CC\\
    L&East&35&5.9&Residential&X&South&34&7.5&To CC\\
    L&West&36&6.0&Residential&Y&North&30&8.0&To CC\\
    M&East&18&4.4&Residential&Y&South&28&8.0&From CC\\
    \bottomrule
\end{tabular}
\end{center}
\end{table}

\begin{figure}[!ht]
    \centering
    \includegraphics[scale=0.5]{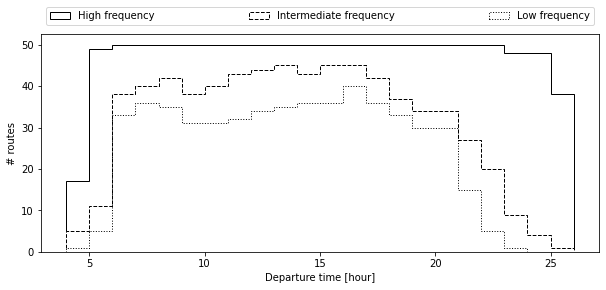}
    \caption{Average route frequency, with frequency defined by high frequency: 6 trips per hour and more; intermediate frequency: between 3 trips and 5 trips per hour; low frequency: less than 3 trips per hour}
    \label{fig:frequency_per_hour}
\end{figure}

The scheduled departure time of a trip is fixed when solving the timetabling or headway setting problem. If the trip is headway-based, the duration between the departure time of two consecutive trips is constant for a given time segment. Otherwise, if the trip is timetabled, the scheduled departure time is established in order to fulfill a given minimum level of service. In general, a stable frequency helps passenger transparency regarding the passing time of buses. In the case of the 50 selected routes, a mix of headway-based and timetabled trips were recorded. Figure \ref{fig:frequency_per_hour} presents the average route frequency breakdown. The frequency is divided in 3 categories: high frequency (6 trips per hour and more), intermediate frequency (between 3 trips and 5 trips per hour) and low frequency (less than 3 trips per hour). For every route, the average frequency per hour is identified and classified in one category. Then, we count the number of routes per category and per hour. In Figure \ref{fig:frequency_per_hour}, we can see that a majority of the selected routes have a high frequency during daytime (6:00AM to 20:59PM). In contrast, the majority of the routes have a lower frequency during the early morning (4:00AM to 5:59AM) and during the end of the day (21:00PM to 25:59PM). The lines O and P are the only ones with a low frequency all day long. 

\subsection{Features analysis}
The PDTT has to be based upon features (i.e., explanatory variables) that are available a few days or weeks in advance. For example, meteorological conditions are likely to influence the TT duration. However, since it is an information that is not available when solving service planning problems, it is not considered. Likewise, the TT of the previous trip is not considered. The list of possible features includes the day of the week, type of region, route identifier, distance, number of stops, scheduled departure time, week number and year. Possible values and types of features (categorical or non-categorical) are listed in Table \ref{table:features_type}.

\begin{table}[h]
\caption {Long-term features}
\label{table:features_type}
\begin{center}
\begin{tabular}{lll}
    \toprule
    Feature&Type&Possible values\\
    \midrule
    Day of the week&Categorical&\{Monday, Tuesday, ..., Friday\}\\
    Region&Categorical&\{residential, crossing CC, ..., to indust.\}\\
    Route identifier&Categorical&\{A East, A West, ..., Y South\}\\
    Distance (km)&Non-categorical&[3, 15.3]\\
    Number of stops&Non-categorical&\{17, 18, \dots, 74\}\\
    Scheduled departure time&Non-categorical&[4:00AM, 25:59PM]\\
    Week number&Non-categorical&\{35, 36, ..., 44\}\\
    Year&Non-categorical&\{2017\}\\
    \bottomrule
\end{tabular}
\end{center}
\end{table}

\begin{figure}[!ht]
    \centering
    \includegraphics[scale=0.5]{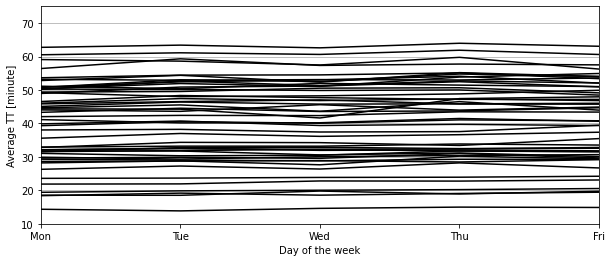}
    \caption{Average TT per day of the week}
    \label{fig:weekday_TT}
\end{figure}

\begin{figure}[!ht]
    \centering
    \includegraphics[scale=0.5]{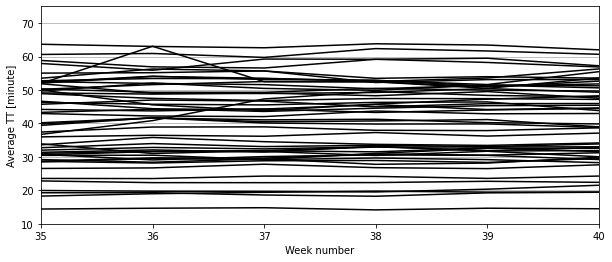}
    \caption{Average TT per week number}
    \label{fig:weekly_TT}
\end{figure}

The feature year is discarded because our dataset is spread over 2017 only. The statistical significance of the features scheduled departure time, day of the week and week number can be analyzed visually by looking at Figures \ref{fig:hourly_TT}, \ref{fig:weekday_TT} and \ref{fig:weekly_TT}, which present the average TT per route depending on each of the feature respectively. Figure \ref{fig:hourly_TT} suggests that the scheduled departure time has a high importance. The relationship between the TT and the scheduled departure time is not linear. Generally, the average TT of a bus route increases during peak hours and is steady between the morning and the afternoon peaks. Second, Figure \ref{fig:weekday_TT} suggests that the relationship between the TT and the day of the week is less important. This is hardly surprising given that Saturdays and Sundays are not considered. Interestingly, there is no common pattern between the routes; for example some routes have a slightly higher average TT on Tuesdays than on Mondays and Wednesdays, while some others have an inverse pattern (i.e., the average TT on Tuesdays is slightly lower than on Mondays and Wednesdays). Third,  Figure \ref{fig:weekly_TT} suggests that the relationship between the TT and the week number is significant only for a handful of bus routes. In Figures \ref{fig:hourly_TT}, \ref{fig:weekday_TT} and \ref{fig:weekly_TT}, we can observe that the average TT differs greatly from one bus route to another, unsurprisingly as each route has its own characteristics (as discussed earlier).

Furthermore, the statistical significance of all the features can be assessed by the permutation feature importance \citep{breiman2001}, which measures the increase of the MSE of a Random Forest estimator when the values of a feature are permuted. The importance of a feature $\ell$ is the difference between a Random Forest model's MSE of the original dataset and the average (over 10 shuffles) MSE of a corrupted dataset (with the values of feature $\ell$ permuted). Results of this analysis are presented in Table \ref{table:features_importance}. It indicates that the two most important features are the number of stops and the distance traveled, which are two main characteristics of a bus route. When the number of stops and the distance traveled is known, the route identifier is less statistically significant.  It also confirms that the scheduled departure time is an important feature to predict the TT. The week number is also significant, but impacts the prediction error to a smaller extend. The day of the week and the type of operational region have a limited impact. 

\begin{table}[!ht]
\caption {Permutation feature importance}
\label{table:features_importance}
\begin{center}
\begin{tabular}{lll}
    \toprule
    Feature&Importance&Relative importance (\%)\\
    \midrule
    Number of stops&185.76&51.45\\
    Distance&62.37&17.28\\
    Scheduled departure time&56.95&15.77\\
    Route identifier&33.73&9.34\\
    Week number&8.51&2.36\\
    Day of the week&7.74&2.14\\
    Region&5.96&1.65\\
    \bottomrule
\end{tabular}
\end{center}
\end{table}

\section{Methodology}

Several approaches have been proposed to compute the conditional p.d.f. of a random variable in a probabilistic fashion. Gaussian process based models are among the most popular. The work of \cite{dutordoir2018} proposed a Gaussian process-based model to estimate a conditional p.d.f. using latent variables in order to model non-Gaussian probability distributions. Also, \cite{Bishop1994} developed Mixture Density Networks, which is a type of artificial neural network predicting multimodal conditional density distributions. The main drawback of Mixture Density Networks is that they perform poorly when the size of the dataset is not large enough. In the work of \cite{Yeo2018}, the prediction of a continuous p.d.f. is converted into a classification task by using a discretization technique. This simplifies the learning task and traditional probabilistic classifiers can be used to predict the probability mass function, which can be smoothed later on into a p.d.f.. The focus of this paper is on frequentist models \citep{Koller2009}. In the remainder of this section two approaches for the PDTT are presented. The first one estimates the p.d.f. of the TT of a set of similar trips using parametric, semi-parametric or non-parametric density estimation models. The second approach, namely the smoothed Logistic Regression for probabilistic classification, is similar to that of \cite{Yeo2018}, but fits a Logistic Regression instead of a Recurrent Neural Network estimator.

\subsection{Similarity-based density estimation}
\label{Sec:similarity-based_density_estimation}
Similarity-based density estimation models are a two-step process: for each trip, (1) find the set of similar trips and (2) estimate the density of this particular set, by fitting a parametric, semi-parametric or non-parametric model. Next, we will define two similarity-based methods and introduce some density estimation models.

\subsubsection{Similarity-based methods}
\label{sec:similarity}
Consider a trip $i$ with a feature vector $(\boldsymbol{x}_1^{(i)}, \dots, \boldsymbol{x}_d^{(i)})$. We want to select, based upon one of the following similarity-based methods, the set of trips in the (reduced) training set that have similar attributes:

\begin{itemize}
    \item Equivalent DTW (eDTW) : select all trips from the same route that have a scheduled departure time in the same departure time window (DTW) \citep{Mazloumi2009, Buchel2018, Ma2016} than trip $i$.

    \item \textit{k} Nearest Neighbors (\textit{k}NN): select the $k$ nearest neighbors of  the trip $i$. The distance between trip $i$ and trip $j$ is the Euclidean distance between their feature vectors and is computed as
    
    \begin{equation}
        dist(\boldsymbol{x}_i,\boldsymbol{x}_j) = \sqrt{\sum_{\ell=1}^d (x_\ell^{(i)} - x_\ell^{(j)})^2}.
    \end{equation}
\end{itemize}

\subsubsection{Density estimation models}
The conditional probability $p(T_i= t_i \mid \boldsymbol{x}_i)$ is estimated by fitting a given density estimation model on points close to $i$ in the (reduced) training set, which are either trips in the same eDTW or close neighbors.\\

\noindent \textit{Parametric models}

Parametric density estimation considers a restricted set of common probability distributions. Each of these distributions has a small number of parameters that have to be estimated from the data. We consider the Normal, Log-Normal, Logistic, Log-Logistic, Gamma, and Cauchy probability distributions. The Gamma distribution is a family of probability distributions containing the Exponential, Erlang and Chi-Squared distributions. In the work of \cite{Ma2016}, the first four distributions were successful at modeling the TT distribution at a route level. For each trip, parameters of these probability distributions are found using the Maximum Likelihood Estimation (MLE) algorithm.\\

\noindent \textit{Semi-parametric model: Gaussian Mixture Model}

GMMs are a sub-category of mixture models composed of $K$ normal components. GMMs are relevant when the population modeled is multimodal and has undefined subpopulations or states, such that each component represents a state. It is common in transport to use three components, one for each of the traffic states: free flow, recurrent and non-recurrent traffic \citep{Ma2016}. The p.d.f. of a $K$-components GMM is given by 

\begin{equation}
\label{eq:PDF_GMM}
    \hat{p}(T_i = t_i \mid \boldsymbol{x}_i) = \sum_{k=1}^{K} \pi_{ik} \mathcal{N}(t_i \mid \mu_{ik}, \sigma^2_{ik}).
\end{equation}

The vector of positively defined coefficients $\mathbf{\pi}_i = (\pi_{i1}, ..., \pi_{iK})$, such that $\sum_{k=1}^{K} \pi_{ik} = 1$, and the vector of the model's $k^{th}$ component's parameters $(\mu_{ik}, \sigma^2_{ik})$, are found by applying the expectation maximization (EM) algorithm.\\

\noindent \textit{Non-parametric model: Kernel Density Estimation (KDE)}

A KDE model infers the p.d.f. of a random variable based on a sample of its population. To estimate the p.d.f. of a trip $i$, the model uses $m$ points close to $i$ in the training set. It is a data smoothing problem that allows to find, in a non-parametric fashion, the curve of the p.d.f. given a sample. The Gaussian kernel, $K(\cdot)$, is the most widely used, but any function that integrates to unity ($\int K(t)dt = 1$) can replace it. The smoothness of the estimator is adjusted by the bandwidth parameter $h$ as
\begin{equation}
    \hat{p}(T_i = t_i \mid \boldsymbol{x}_i)  = \frac{1}{m}\sum_{j = 1}^{m} K \left(\frac{t_i - t_j}{h} \right).
\end{equation}

\subsection{Smoothed Logistic Regression for probabilistic classification (LR-PC)}
Probabilistic classifiers are a type of machine learning model that can predict the probability that a given input belongs to a set of classes, instead of only predicting the class with the highest probability. When a numerical discretization is applied to the random variable $T$, such that the TT is categorized in bins of 1 minute, the PDTT task can be translated into a probabilistic classification one: estimate the probability that $T$ takes a value that falls into class $c \in \{0, ...,C - 1\}$. A model is fitted per bus route, because this setting yields better experimental results than learning a unique model for all bus routes (see Section \ref{sec:param_selection}).

A question arises: how to choose the number of classes for a bus route. Our approach was to use, for a given bus route, the difference between the trip in the training set with the shortest duration, $t_{min}$, and the trip with the longest duration, $t_{max}$, as $C$, the number of classes. Thus, $P(T_i = c)$ is the probability that $T_i$ takes a value in $[c + t_{min} , c+1 + t_{min}[$. We disregarded the fact that trips in the test set can have shorter or longer duration than $t_{min}$ and $t_{max}$ respectively, as the smoothing discussed later implicitly solves this issue.

Multinominal Logistic Regression is naturally probabilistic and is commonly used for probabilistic classification tasks. Classes' probabilities of a multinomial Logistic Regression are defined as 

\begin{equation}
    \hat{P}(\floor*{T_i} = c + t_{min} \mid \boldsymbol{x}_i, \mathbf{w}_c) = \frac{exp(\mathbf{w}_c^T \boldsymbol{x}_i)}{\sum_{c^\prime = 0}^{C -1}exp(\mathbf{w}_{c^\prime}^T \mathbf{x}_i)},
\end{equation}

\noindent where $\mathbf{w}_c$ is the vector of parameters of the class $c$, found using a stochastic average gradient descent solver.

Logistic Regression outputs a probability mass function (p.m.f.) that can be smoothed into a p.d.f. subsequently. As proposed by \cite{Yeo2018}, the output of the multinomial Logistic Regression, which takes the form of a p.m.f., is passed thought a one-dimensional convolution layer. This step has the effect of enforcing a spatial correlation in the output. The convolution layer is analogous to a KDE with a bandwidth $h$ and a kernel $K(\cdot)$, but uses the p.m.f, $\hat{P}(\floor*{T_i} \mid \boldsymbol{x}_i, \boldsymbol{w}_c)$, instead of a sample of trips:

\begin{equation}
    \hat{p}(T_i = t_i \mid \boldsymbol{x}_i) = \sum_{c' = 0}^{C-1} \left[K \left(\frac{t_i - (c' + t_{min})}{h}\right) \times \hat{P}(\floor*{T_i} = c' + t_{min}\mid \boldsymbol{x}_i, \mathbf{w}_{c'}) \right].
\end{equation}

\section{Experimental results}

Using the dataset described in Section \ref{sec:data}, we next explain how we fit probabilistic models for the PDTT, before comparing the performance of these models. First, we describe how the data is filtered and split for the PDTT. Second, we go through features and parameters selection for each type of model and detail the selection of the temporal aggregation level. Finally, all probabilistic models are compared to a Random Forest model in terms of their performance on the test set.

\subsection{Data preparation}
The dataset is filtered in order to remove erroneous information and special situations that we do not want to cover in the PDTT. Incomplete trips, vias and trips with detours are discarded. A via is a trip that deviates from the main trip or an express trip. To remove erroneous trips which were not filtered by the previous step, the Median Absolute Deviation or MAD \citep{Hellerstein2008} with a 6-delta criterion is used. A trip is discarded if it has a TT longer or shorter than the median TT value of the trips associated with the same route plus (minus) 6 times the corresponding standard deviation. This method also removes trips with extended TT due to exceptional scenarios (e.g., bus failure) that the PDTT problem should not cover, because when such exceptional scenarios occur, an additional bus is usually dispatched to recover the schedule and prevent severe delay propagation.

Figure \ref{fig:dataset_split} summarizes the dataset split. We split the dataset such that the training data starts on 08/28/2017 and ends on 10/08/2017. The set of test data is composed of the trips from 10/16/2017 to 10/29/2017. Hence, a complete week, from 10/09/2017 to 10/16/2017 is discarded to simulate real-life settings where the planning is done at least a few days ahead. The training set is split again in a validation set and a reduced training set by slicing the last  trips recorded per route from the original training set. The reduced training set contains 80\% of the trips in the original training set and the validation set contains the remaining 20\%. The validation set is used for features and parameters selection, which is detailed in the next section. 

\begin{figure}[h!]
\centering
\caption{Dataset split}
\label{fig:dataset_split}
\begin{tikzpicture}[%
    every node/.style={
        font=\scriptsize,
        text height=1ex,
        text depth=.25ex,    
    },
]
\clip (-2,-1) rectangle (10, 1);
\draw[->] (0,0) -- (9.5,0);

\foreach \x in {0,1,...,9}{
    \draw (\x cm,3pt) -- (\x cm,0pt);
}

\node[anchor=north] at (0,0) {$08/28$};
\node[anchor=north] at (5,0) {$10/08$};
\node[anchor=north] at (6,0) {$10/16$};
\node[anchor=north] at (8,0) {$10/29$};
\node[anchor=north] at (9.5,0) {date};

\fill[myLightGray] (0,0.25) rectangle (5,0.4);
\fill[pattern=north east lines] (3.8,0.25) rectangle (5,0.4);

\fill[myLightGray] (6,-0.4) rectangle (8,-0.55);

\node[anchor=west] at (-2,0.35) {Training set};
\node[anchor=west] at (-2,-0.35) {Test set};

\draw[decorate,decoration={brace,amplitude=5pt}] (3.8,0.45) -- (5,0.45)
    node[anchor=south,midway,above=4pt] {Validation set};
\end{tikzpicture}
\end{figure}

\subsection{Features and parameters selection}
\label{sec:param_selection}

Similarity-based density estimation models use one of the similarity-based methods, namely the eDTW or the $k$NN method. While the eDTW method does not require feature selection, as the features used are always the route identifier and the scheduled departure time, the $k$NN method does require feature selection. Indeed, the distance between neighbors depends on the specified feature vector. Estimating the TT density of a trip $i$ does not require feature selection; it fits a probability density to a sample containing trips similar to trip $i$. Thus, here the features selection problem is reduced to finding a feature vector for the $k$NN method. Features considered are the ones listed in Table \ref{table:features_importance}. The selection of features is carried out in parallel with the selection of parameters; we are looking for the combination of features and parameters that yields the best NLL of the validation set. The parameters of the similarity-based methods are the DTW duration for the eDTW method and the number $k$ of neighbors for the $k$NN method. For the KDE, the validation set is used to select the bandwidth $h$ and the kernel function.

We found that similarity-based density estimation models fit the data better when a $k$NN is fitted per bus route. By doing so, the features describing the bus route characteristics, namely the number of stops, distance traveled, route identifier and type of region, become uninformative to the model. Indeed, all trips used to train the model of a given bus route have exactly the same values for these features. The remaining features to consider are the scheduled departure time, the week number and the day of the week. For the $k$NN method, a feature vector composed of the two former features gives the best results. Thus, the closest neighbors of a trip $i$ are defined as trips of the same bus route with a scheduled departure and a week number in a temporal horizon close to trip $i$. Temporal aggregation is a fundamental aspect of the PDTT since it has been shown to affect the shape and nature of the TT probability distribution \citep{Mazloumi2009, Ma2016}. Thus, the parameter associated to it, namely the DTW duration, is studied carefully.

\begin{table}[ht!]
\caption {NLL (lower is better) of the similarity-based density estimation models using eDTW method with different levels of temporal aggregation}
\label{table:NLPP_aggre}
\begin{center}
\begin{tabular}[t]{lrrrrrrrrr}
    \toprule
    &&\multicolumn{2}{c}{5 periods}&&\multicolumn{2}{c}{60 minutes}&&\multicolumn{2}{c}{30 minutes}\\
    \cline{3-4}\cline{6-7}\cline{9-10}
    Model&&Train&Validation&&Train&Validation&&Train&Validation\\
    \midrule
    Cauchy&&3.24&3.30&&2.72&2.89&&2.60&2.96\\
    Gamma&&3.08&\textbf{3.15}&&2.58&2.83&&2.47&5.98\\
    Normal&&3.09&3.16&&2.59&2.87&&2.48&6.57\\
    Log-Normal&&3.09&\textbf{3.15}&&2.57&2.81&&2.46&5.72\\
    Logistic&&3.09&3.16&&2.59&2.77&&2.48&2.91\\
    Log-Logistic&&3.09&\textbf{3.15}&&2.58&\textbf{2.75}&&2.47&\textbf{2.88}\\
    GMM&&3.07&3.16&&2.26&5.81&&1.82&24.96\\
    KDE&&2.99&\textbf{3.15}&&2.49&2.76&&2.77&2.89\\
    \bottomrule
\end{tabular}
\end{center}
\end{table}

We select the  DTW duration by analyzing how the density estimation models perform on the validation set for different levels of temporal aggregation. DTWs considered are, going from the most aggregated to the least aggregated, 5 periods per day (before morning peak, morning peak, in-between morning and afternoon peaks, afternoon peak and after afternoon peak), 60 minutes and 30 minutes. Table \ref{table:NLPP_aggre} compares the performance of models using the eDTW method, with the values in bold indicating the best NNL of the validation set for each level of temporal aggregation. For all models except the GMM, the NLL score of the validation set is better at DTWs of 60 minutes. The Log-Logistic model obtains the best results at all aggregation levels, matched by the Gamma, Log-Normal and KDE models at an aggregation level of 5 periods per day. The GMM has a similar performance to the parametric models for the most aggregated level, but it also has a poor performance for lower levels of temporal aggregation, both 60 and 30 minutes. Since the performance on the reduced training set is good, it indicates that the GMM overfits the training data. For the DTWs considered, the conditional p.d.f. of the TT is most likely not multimodal. Thus, this model is discarded for the rest of the study. Interestingly, the training NLL decreases when the temporal aggregation level increases for all models except the KDE, while the NLL of the validation set increases from DTWs of 60 minutes to DTWs of 30 minutes for all models. This suggests that models are overfitting more at DTWs of 30 minutes than at DTWs of 60 minutes. Indeed, when the level of temporal aggregation decreases, the DTW duration and thus the number of training points per DTW decreases, which can result in a predicted conditional p.d.f. that does not fit the validation set points.

\begin{figure}[ht!]
    \centering
    
    \begin{subfigure}{5cm}
    \centering\includegraphics[width=5cm]{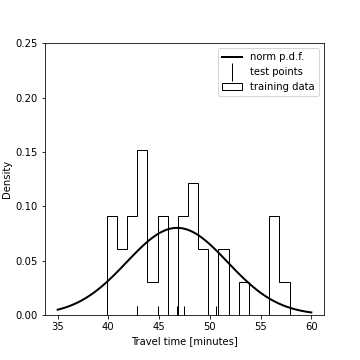}
    \caption{9:00 to 10:00}
    \label{fig:60min}
    \end{subfigure}
    \begin{subfigure}{5cm}
    \centering\includegraphics[width=5cm]{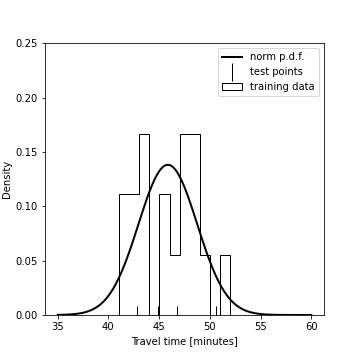}
    \caption{9:00 to 9:30}
    \label{fig:30min_p1}
    \end{subfigure}
    \begin{subfigure}{5cm}
    \centering\includegraphics[width=5cm]{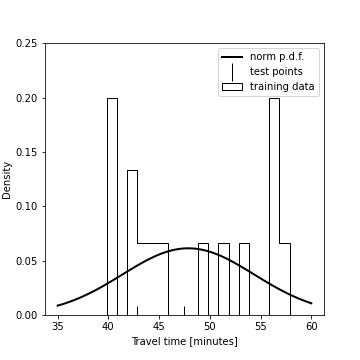}
    \caption{9:30 to 10:00}
    \label{fig:30min_p2}
    \end{subfigure}
    
    \caption{Example of the fit of a Normal distribution with (a) 60 minutes and (b), (c) 30 minutes DTWs, route H-South.}
    \label{fig:aggre_level_30_vs_1h}

\end{figure}

Figure \ref{fig:aggre_level_30_vs_1h} illustrates an overfitting case, where Figure \ref{fig:60min} is the predicted p.d.f. of the TT of a trip when the DTW used for the trip selection is 60 minutes and Figures \ref{fig:30min_p1} and \ref{fig:30min_p2} are the corresponding p.d.f.s of the TT for DTWs of 30 minutes. Indeed, when the set of points between 9:00AM and 10:00AM is divided in two DTWs of 30 minutes, the second portion (Figure \ref{fig:30min_p2}) is flat with two peaks at the ends. The model estimation of the p.d.f. of the TT is poor. This is probably due to the small number of data collected between 9:30AM and 10:00AM, which results in an over representation of the tails of the probability distribution.

\begin{figure}[]
    \centering
    \includegraphics[scale=0.5]{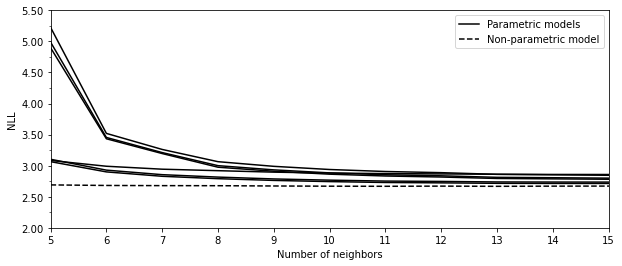}
    \caption{NLL (lower is better) of the similarity-based density estimation models using $k$NN method with different numbers of neighbors}
    \label{fig:nb_neighbors}
\end{figure}

We can conclude that, between the three levels of temporal aggregation compared, the best one is the one with DTWs of 60 minutes. We denote the eDTW  method with DTWs of 60 minutes as eDTW*. For the second similarity-based method, namely the $k$NN, the value of $k$ can be chosen similarly to the duration of DTWs, by assessing the performance of the similarity-based density estimation models on the validation set for different values of $k$. Figure \ref{fig:nb_neighbors} shows that the NLL of the validation set decreases significantly  for all models when the value of $k$ increases up to approximately $k = 13$. After that, the NLL stays almost constant. Thus, we set the number $k$ of neighbors to 13 and denote the $k$NN method with $k = 13$ as $k$NN*.

The LR-PC model also yields better performance when it is fitted per bus route and when it has a feature vector containing only the scheduled departure time and some transformations thereof. The scheduled departure time is transformed in order to capture a non-linear relationship with the TT. First, it is categorized in bins of 1 hour and 30 minutes using a one-hot encoding. Second, the sine and cosine of the scheduled departure time are computed. The total dimension of the feature vector is 69 (22 for the one-hot encoding of bins of 1 hour, 44 for the one-hot encoding of bins of 30 minutes, 2 for the sine and cosine of the scheduled departure time and 1 for the original scheduled departure time). The bandwidth and the kernel function of the LR-PC model are selected based on the performance on the validation set, along with the regularization strength of the Logistic Regression.

The Random Forest model's NLL of the validation set is minimized with a feature vector composed of the number of stops, distance traveled, scheduled departure time, route identifier, week number and day of the week. Thus, only the feature encoding the type of region is discarded. This is in line with the results of the permutation feature importance (see Table \ref{table:features_importance}). The number of trees in the Random Forest model, the maximum number of features considered when branching, the maximum depth of each tree and the minimum number of samples required to split an internal node are selected based on the NLL of the validation set.

\subsection{Models comparison}
\label{sec:model_compa}
Table \ref{table:NLPP_allmodels} presents the NLL of the training set (including the validation set) and of the test set for all models. For the parametric models, the Cauchy, Logistic and Log-Logistic distributions have better results on the test set when using the $k$NN* method, while the Gamma, Normal and Log-Normal have rather the opposite results. The NLL of the training set is smaller for all models when the $k$NN* method is used. This indicates that the $k$NN* method fits better the training points than the eDTW* method, but may also indicates a mild overfitting (especially for Gamma, Normal and Log-Normal distributions which have a better test NLL when applied with the eDTW* method). The KDE model performs better when it uses the $k$NN* method than when it uses the eDTW* method. Overall, the KDE model using $k$NN* method is the one which yields the best result on the test set, closely followed by the Log-Logistic with $k$NN* and the LR-PC models.

\begin{table}[ht!]
\caption {NLL (lower is better) of the training and test sets of similarity-based density estimation, LR-PC and Random Forest models}
\label{table:NLPP_allmodels}
\begin{center}
\begin{tabular}[t]{llrrr}
    \toprule
    Model&similarity&Training&Test\\
    &method&set&set\\
    \midrule
    Cauchy&eDTW$^*$&2.73&2.87\\
    &\textit{k}NN$^*$&2.65&2.84\\
    Gamma&eDTW$^*$&2.59&2.79\\
    &\textit{k}NN$^*$&2.51&2.80\\
    Normal&eDTW$^*$&2.61&2.82\\
    &\textit{k}NN$^*$&2.52&2.84\\
    Log-Normal&eDTW$^*$&2.59&2.77\\
    &\textit{k}NN$^*$&2.51&2.78\\
    Logistic&eDTW$^*$&2.61&2.74\\
    &\textit{k}NN$^*$&2.53&2.72\\
    Log-Logistic&eDTW$^*$&2.59&2.72\\
    &\textit{k}NN$^*$&2.52&2.70\\
    KDE&eDTW$^*$&2.61&2.73\\
    &\textit{k}NN$^*$&2.41&\textbf{2.69}\\
    LR-PC&-&2.48&2.70\\
    Random Forest&-&2.45&2.80\\
    \bottomrule
\end{tabular}
\end{center}
\smallskip
\parbox[t]{\textwidth}{\footnotesize
$^*$DTWs = 60 minutes or \textit{k} = 13
}
\end{table}

Similarity-based density estimation models and the LR-PC model are compared with a Random Forest model. As shown in Table \ref{table:NLPP_allmodels}, the Random Forest is outperformed by all the models except the Gamma distribution using $k$NN* method as well as the Normal and the Cauchy distributions using eDTW* or $k$NN* method, as it has a larger test NLL. A Random Forest is not inherently probabilistic; it is usually used for \textit{point prediction} \citep{dutordoir2018}.  Indeed, the emphasis is on modeling the mapping between an input $\boldsymbol{x}$ to its output $y$ rather than on predicting the conditional p.d.f. $p(y \mid \boldsymbol{x})$ \citep{dutordoir2018}. When optimized with a sum-of-squares function, a Random Forest model can be interpreted as fitting a Gaussian on the conditional distribution of the TT, where the mean of the Gaussian is given by the output of the Random Forest and the variance is constant for all the inputs and equal to the MSE of the output. Thus, these experimental results show that there is an added value in modeling the conditional p.d.f. of the TT using probabilistic models. A key factor could be that in probabilistic models the variance is not constant for every trip.  It is interesting to see that models using a non-Gaussian probability distribution, either a Gamma, Log-Normal, Logistic or Log-Logistic distribution outperform the one using a Normal distribution, which calls the normality of the conditional p.d.f. of the TT into question. 

The PDTT objective is to predict the complete conditional p.d.f. of the TT given a trip's feature vector. However, we can assess the performance of the models developed for the PDTT for a different type of prediction: \textit{point prediction} \citep{dutordoir2018}, which aims at predicting the expected value of the TT of a trip. A Random Forest optimized with a sum-of-squares function is fitted in order to minimize the MSE of the difference between the prediction (i.e., the expected value of the TT) and the true TT duration. We compare the \textit{point prediction} performance of the Random Forest model to the performance of the models yielding the best NLL score, namely the Log-Logistic model using $k$NN* method, the KDE model using $k$NN* method and the LR-PC model. Table \ref{table:MSE} shows that these all have an MSE score of the same order of magnitude as the Random Forest model. Furthermore, after the Random forest model, the KDE model using $k$NN* method has the lowest MSE. Thus, the KDE model using $k$NN* predicts a complete conditional p.d.f. that corresponds better to the true region in which a trip duration is likely to lie than the Random Forest model and in addition it is able to compute \textit{point predictions} almost as accurately as the Random Forest model (less than 3\% increase in MSE). 

\begin{table}[ht!]
\caption {MSE of the TT on the test set}
\label{table:MSE}
\begin{center}
\begin{tabular}[t]{llr}
    \toprule
    Model&similarity&MSE\\
    &method&\\
    \midrule
    Log-Logistic&\textit{k}NN$^*$&16.23\\
    KDE&\textit{k}NN$^*$&16.18\\
    LR-PC&-&16.94\\
    Random Forest&-&\textbf{15.71}\\
    \bottomrule
\end{tabular}
\end{center}
\smallskip
\parbox[t]{\textwidth}{\footnotesize
$^*$\textit{k} = 13
}
\end{table}

\section{Conclusions}
In public transport, reliability has become a key challenge for operators wishing to attract new users. We proposed to address the reliability of bus services at the planning stage, and more precisely when computing one of the problem of service planning, namely the vehicle scheduling problem. To that end, we defined the reliable MDVSP with stochastic TTs, a version of the vehicle scheduling problem taking in input a set of timetabled trips and the conditional p.d.f. of their TT in order to output reliable vehicle schedules (i.e., vehicle schedules with small expected secondary delays for all timetabled trips). The prediction of the p.d.f. of the TT, that we referred to as the PDTT, is done a few days ahead of operations in order to incorporate the prediction in the planning phase.  We investigated if probabilistic models can predict more accurately the complete conditional p.d.f. of the TT than a Random Forest model. The latter is not inherently probabilistic, as it is typically used for \textit{point prediction}, but under some conditions it can be interpreted as fitting a Gaussian on the conditional distribution of the TT given the characteristics of a set of trips.

We used a 2-month dataset collected by buses equipped with APTS in the city of Montréal, involving 50 bus routes and a total of over 41,000 trips. The bus routes studied have various attributes (e.g., number of stops, frequency, traveled distance, etc.) and constitute a diverse sample from which we hope to obtain results transferable to other bus networks. Based on previous works on TT variability analysis, we determined a set of features that we ranked in order of decreasing significance as follows: number of stops, distance, scheduled departure time, route identifier, week number, day of the week, type of region.

In order to compute the conditional p.d.f of the TT, we proposed two types of probabilistic models, namely similarity-based density estimation models and the LR-PC model. The former is a two-step process that firstly find, for each trip, the set of similar trips and then estimate the density of this set using parametric, semi-parametric or non-parametric density estimation models. We proposed two types of similarity-based methods, namely the eDTW and the $k$NN, for which the temporal aggregation level and the number of neighbors had to be set, respectively. The LR-PC model applies a numerical discretization to the TT before fitting a Logistic Regression classifier per bus route. The output of the Logistic Regression is then smoothed into a p.d.f. using a convolution layer analogous to a KDE.

Previous works on TT distribution modeling indicated that the level of temporal aggregation greatly affects the shape and nature of the TT distribution. Thus, we carefully selected the DTW duration based on the performance on the validation set. The GMM model had a poor performance for DTWs of 60 minutes and 30 minutes and thus we concluded that the conditional p.d.f. of the TT is most likely not multimodal. This result is aligned with the one of \citet{Ma2016} which observed that the multimodality of the TT decreases with spatial aggregation. Finally, models were compared in terms of their test NLL. From all the models tested, the density-based estimation model using $k$NN method and a KDE yielded the best NLL score. The Random Forest model was outperformed by most of probabilistic models and had comparable \textit{point prediction} accuracy (measured by the MSE of the TTs) to models with the best NLL scores. This result indicated that there is an added value in modeling the conditional p.d.f. of the TT using probabilistic models. In particular, probabilistic models allow the variance of the p.d.f. of the TT to be different for every trip, as opposed to a Random Forest model as we interpreted it. Also, as similarity-based density estimation models using the Normal distribution were outperformed by several other parametric distributions, the normality of TTs was questioned, supporting the benefit of using probabilistic models for the PDTT. 

\section*{Acknowledgements}
The authors would like to thank Charles Fleurent and his team at GIRO Inc. for their valuable help and contributions and the Société de transport de Montréal (STM) for sharing the dataset. The advice of Didier Chételat and Maxime Gasse are greatly appreciated. We gratefully acknowledge the financial support provided by GIRO Inc. and the Natural Sciences and Engineering Research Council of Canada under the grant CRDPJ 520349-17 as well as the financial support provided by the Institute for Data Valorization (IVADO).

\bibliographystyle{bib_style}
\bibliography{refs}

\end{document}